\documentclass[letterpaper, 10 pt, conference]{ieeeconf}  

\IEEEoverridecommandlockouts                              

\usepackage{amssymb}
\usepackage{lscape}
\usepackage{multirow}
\usepackage{algorithm}
\usepackage{algpseudocode}
\usepackage{amsmath} 
\usepackage{hyperref}
\usepackage{dsfont}
\usepackage{minted}
\usepackage{bbm}
\usepackage{listings}
\usepackage{graphicx}
\usepackage[compatibility=false]{subcaption}
\usepackage{booktabs}
\usepackage{arydshln}
\usepackage{wrapfig}
\usepackage{makecell}
\newenvironment{fullwidthquote}
  {\begin{list}{}%
    {\leftmargin=4pt
     \rightmargin=0pt
     \topsep=4pt
     \parsep=0pt
     \itemsep=0pt}%
   \item\relax}
  {\end{list}}

\title{\LARGE \bf
MAGNIFIED: RL Fine-tuning of Multimodal Large Language Models for Motion Planning
}

\author{Letian Chen$^{*1}$, Yiren Lu$^{*1}$, Justin Fu$^{1}$, Yichen Xie$^{1}$, Runsheng Xu$^{1}$, Jyh-Jing Hwang$^{1}$, \\Ben Sapp$^{1}$, Drago Anguelov$^{1}$
\thanks{*Equal Contributions. $^{1}$Waymo LLC.}
}

\begin{document}

\maketitle
\thispagestyle{empty}
\pagestyle{empty}

\begin{abstract}
Multi-modal Large Language Models (MLLMs) have demonstrated remarkable capabilities in semantic understanding and common sense reasoning, making them promising candidates for solving planning problems in autonomous driving. However, the next-token text prediction objectives traditionally used in pre-training and supervised fine-tuning (SFT) of MLLMs may fall short of fulfilling the planning objectives for autonomous vehicles. The next-token prediction objective merely encourages per-token imitation in text, often irrespective of multi-step consequences and the alignment with crucial planning considerations such as giving space to other road actors. To overcome these limitations, we propose a reinforcement learning fine-tuning (RLFT) approach, MAGNIFIED, that aligns the MLLM-based driving agent with planning objectives by learning from token-level rewards. By mapping a sequence of predicted tokens to corresponding vehicle trajectories and learning from planning rewards, MAGNIFIED optimizes for the true planning objectives rather than focusing solely on token prediction accuracy, enabling the model to refine its understanding of the planning task beyond simple imitation. We validate our approach on the Waymo Open Motion Dataset with a novel setup incorporating rasterized birds-eye views and tokenized trajectories as inputs and planning-oriented outputs. An initial SFT phase establishes a strong baseline in outputting plan trajectories as sequences of X-Y coordinates in text, while subsequent RL fine-tuning substantially enhances planning performance relative to the SFT baseline (demonstrating over a $10.5\%$ reduction in overlap rate and a $38.9\%$ reduction in off-road rate), underscoring the potential of RLFT on MLLMs to achieve vehicle planning that is better aligned with compliant, comfortable, and efficient driving.
\end{abstract}




\section{Introduction}
\label{sec:intro}

Multimodal Large Language Models (MLLMs) have demonstrated significant advancements in tasks that require multi-modal (e.g. visual) comprehension and semantic understanding. Leveraging their ability to process both visual and textual inputs, MLLMs can be adapted through Supervised Fine-Tuning (SFT) for a wide range of applications, including visual question answering~\cite{naseem2022vision}, video captioning~\cite{yang2023vid2seq}, medical image analysis~\cite{bazi2023vision}, robotics~\cite{gao2024physically, xu2023vision}, and autonomous vehicle (AV) perception~\cite{zhou2024vision}. While MLLMs excel in tasks with well-defined answers, their application to AV planning—a domain that demands an understanding of complex scenes, coherent trajectory planning, and decision-making in uncertain environments—remains under-explored~\cite{hwang2024emma}.

AV planning introduces unique challenges that go beyond traditional visual comprehension tasks. Unlike perception, planning requires generating structured sequences that reflect task-specific constraints and preferences. Despite the potential of MLLMs, the standard SFT paradigm of optimizing next-token prediction may fall short on such tasks, as it only encourages imitation of surface-level token distributions observed in the training data, without regard for whether the resulting trajectories are physically feasible, comfortable, or efficient. This gap between token-level prediction and trajectory-level evaluation mirrors challenges in other domains, such as code generation, where producing executable code cannot be ensured through next-token accuracy alone~\cite{dou2024stepcoder}. In contrast to SFT, Reinforcement Learning (RL) is well-suited towards optimizing task-specific objectives, and has been shown to be effective in guiding models toward performance goals~\cite{lu2023imitation, peng2025improving}. 
However, RL training often suffers from high sample complexity, limiting its practical applicability in real-world AV planning, where collecting extensive high-quality data and constructing high-fidelity simulators are both challenging and costly.


\begin{figure*}[t]
    \centering
    \includegraphics[width=0.88\textwidth]{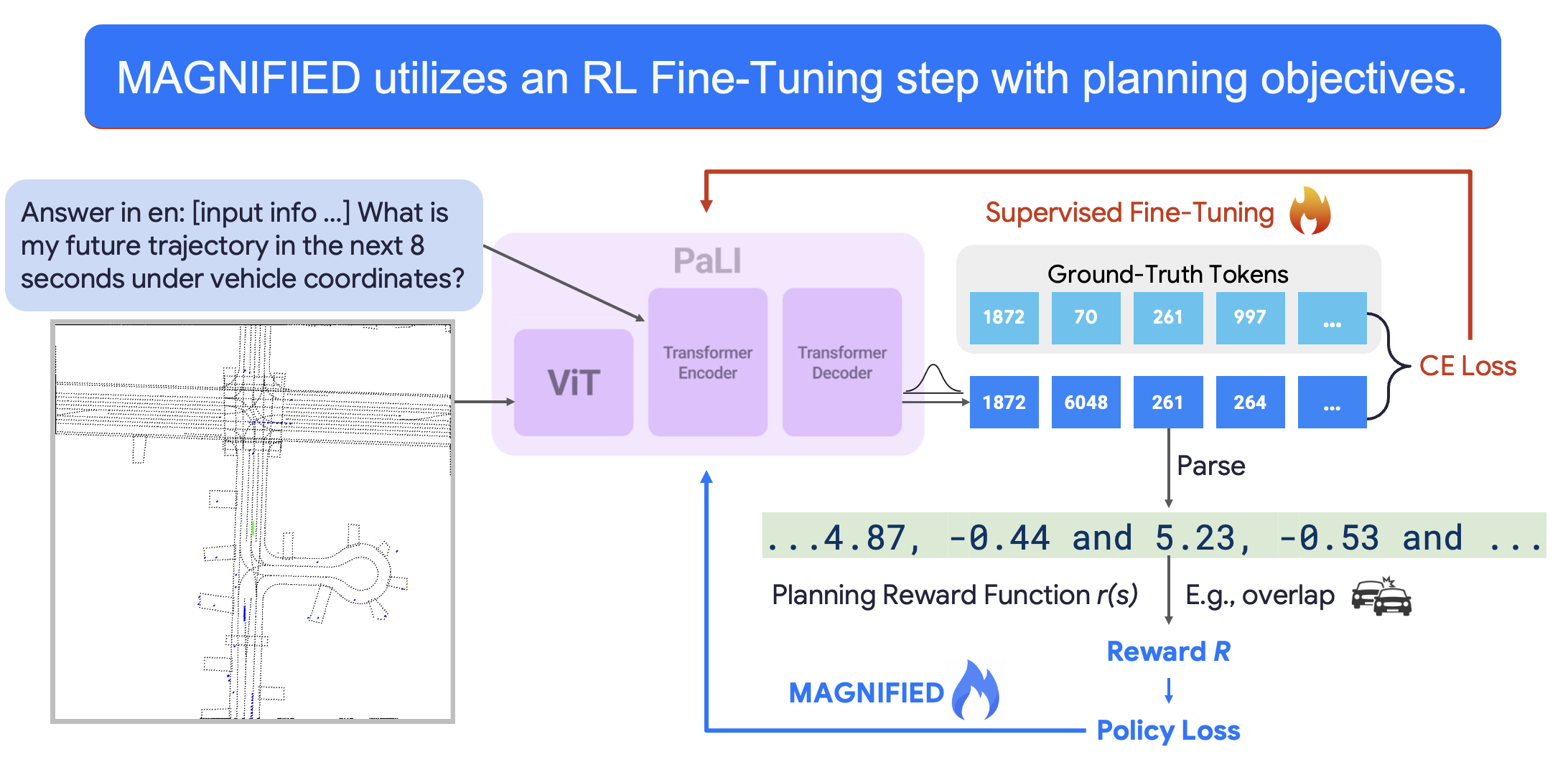}
    \caption{This diagram illustrates our framework for AV planning. Textual and visual inputs are provided to a MLLM, which generates tokens representing the planned trajectory. Traditional SFT utilizes cross-entropy (CE) losses to optimize next-token prediction. In contrast, MAGNIFIED leverages the semantics of the planned trajectory, calculates planning costs, and applies Reinforcement Learning Fine-Tuning (RLFT) to directly optimize for planning cost.}
    \label{fig:overview}
    \vspace{-10pt}
\end{figure*}


To bridge these gaps, we propose a novel approach, MAGNIFIED (reinforceMent leArninG tuNIng For multImodal large languagE moDels), which marries the semantic understanding and common-sense reasoning capabilities of MLLMs with the policy optimization strengths of RL to meet the challenging demands of AV planning. MAGNIFIED leverages the ability of an MLLM base model to interpret a wide distribution of scenarios from extensive pre-training, establishes a strong baseline with planned trajectories as text sequences of X-Y coordinates via a supervised fine-tuning (SFT) phase, and enhances the ability to produce cost-aware plans via an
RL Fine-Tuning (RLFT) phase. The SFT phase vastly reduces the exploration space for the RLFT stage, allowing effective targeted improvements in planning objectives with a relatively small amount of compute (1.25\% of SFT training steps).
MAGNIFIED transforms the MLLM’s output text token sequence into a trajectory representation, enabling a planning reward calculation based on trajectory quality on a per-token basis. Another key advantage of MAGNIFIED lies in its compatibility with non-differentiable rewards, which allows it to optimize arbitrary task-specific rewards that improve final end-to-end planning performance.
Our contributions are threefold: 
\vspace{-5pt}
\begin{enumerate}
    \item We introduce MAGNIFIED, an RLFT method designed to enhance MLLMs for AV planning. MAGNIFIED aligns predicted tokens with planning goals by learning from per-token rewards.
    \item We apply MAGNIFIED on Waymo Open Motion Dataset (WOMD) with a novel setup for planning: rasterizing birds-eye view (BEV) roadgraph and road-user information into images, tokenizing trajectories as inputs, and outputting planned trajectories. 
    \item We demonstrate that MAGNIFIED significantly improves multiple planning objectives compared to SFT: \emph{overlap rate by $10.5\%$}, and \emph{off-road rate by $38.9\%$}, while maintaining or \emph{improving} the imitative metrics, particularly in long-horizon planning, even \emph{without} optimizing imitative objectives, transforming MLLMs from imitative predictors into cost-aware planners.
\end{enumerate}


\section{Related Work}
\label{sec:related_work}
\subsection{MLLMs for Autonomous Driving}
MLLMs have been increasingly explored in autonomous driving to enhance scene understanding and decision-making by integrating visual and textual data. \cite{zhou2024vision} provide a comprehensive survey on MLLM applications in autonomous driving, highlighting their potential in perception, navigation, and control tasks. For example, DriveGPT4~\cite{xu2024drivegpt4}, LMDrive~\cite{shao2024lmdrive}, Drive Anywhere~\cite{wang2024drive}, S4-Driver~\cite{xie2025s4} and OmniDrive~\cite{wang2024omnidrive} utilize LLMs to explain vehicle actions for reasoning and planning. 
Other recent approaches propose a specialized network structure for detailed contextual understanding~\cite{pan2024vlp}, or apply chain-of-thought reasoning (\cite{tian2024drivevlm,wang2024drivecot,bhattacharyya2023look}). Alpamayo-R1~\cite{ullrich2026toward} introduces a structured Chain-of-Causation reasoning framework combined with RL post-training to improve driving reasoning. EMMA~\cite{hwang2024emma} presents promising end-to-end motion planning results with SFT. In contrast, our approach, MAGNIFIED, extends the application of MLLMs to AV planning by utilizing RLFT to optimize true planning objectives, addressing the gap in current research work.
\subsection{RL for Autonomous Driving} 
RL has been widely applied to autonomous driving, focusing on decision-making and control. \cite{kiran2021deep} survey deep RL algorithms for autonomous driving, highlighting challenges such as sample efficiency and the difficulty of collecting datasets that encompass all driving conditions. To address these issues, \cite{wang2023efficient} and \cite{huang2022efficient} improve sample efficiency by incorporating parameterized skills and imitative expert priors, respectively. 
However, these approaches often rely heavily on large datasets and carefully designed rewards. \cite{he2023fear} proposes a neuro-inspired RL framework to enhance safety but notes persistent challenges in generalizing to unseen scenarios. The generalization capabilities of MLLMs, stemming from diverse pre-training data across domains, offer a promising complement to RL. Our method combines the generalization capabilities of MLLMs with the optimization strengths of RL, enabling a more cost-aware planning system. 
\vspace{-5pt}
\subsection{RL Fine-Tuning for LLMs}
RLFT has been explored across various domains to enhance LLMs by aligning their outputs with specific objectives. For instance, \cite{stiennon2020learning} use RL from human feedback to fine-tune language models, improving their alignment with human preferences. \cite{wang2024aligning} explore RLFT to mitigate harmful outputs, demonstrating the effectiveness of RL in refining model behaviors. In the context of vision-language tasks, \cite{ziegler2019fine} apply RLFT to improve image captioning models by improving generated captions on human evaluations. \cite{zhai2024fine} applies RLFT on vision-language models in order to solve a variety of simulated games. Rather than fine-tuning a model directly, \cite{wang2024rl} utilize a MLLM to generate a reward model that can be optimized using RL. 
Despite these advancements, most applications focus on tasks with well-defined outputs, such as image captions or executable codes. In contrast, AV planning presents uncertain, sequential decision-making challenges.
MAGNIFIED extends RLFT to MLLMs in this domain by introducing token-level rewards, enabling precise credit assignment to optimize planning objectives.
\vspace{-5pt}
\section{Preliminaries}
\label{sec:preliminary}

\subsection{Markov Decision Process}
A Markov Decision Process (MDP) is defined by the tuple $(\mathcal{S}, \mathcal{A}, P, R, \gamma)$. $\mathcal{S}$ is the set of states, and $\mathcal{A}$ is the set of actions the agent can take. The transition probability $P(s'|s, a)$ represents the likelihood of reaching a state $s'$ from state $s$ when action $a$ is taken. The reward function $r(s)$ provides a scalar feedback signal at each state, and $\gamma \in [0, 1]$ is the temporal discount factor. 
The goal in RL is to learn a policy $\pi$ that maximizes the expected return, $V(s_t) = \mathbb{E}[\sum_{k=0}^{\infty} \gamma^k r(s_{t+k})]$, starting from state $s_t$. In AV planning, this translates to training a policy that plans future waypoints as actions based on the current state of the environment.
\subsection{REINFORCE with KL Penalty}
The REINFORCE algorithm is a classic policy-gradient method that learns a policy by maximizing the expected return via stochastic gradient ascent~\cite{williams1992simple}, as shown in Equation~\ref{eq:reinforce}. 
\begin{align} 
\label{eq:reinforce}
\max_{\pi_\theta}\mathbb{E}_{(s_t, a_t)\sim \pi_\theta} \left[ \sum_{t=0}^{T} \log \pi_\theta(a_t | s_t) \tilde{Q}_{\pi_\theta}(s_t, a_t) \right]
\end{align}
The state-action value function, $Q$, is typically estimated using the empirical returns, $\tilde{Q}_\pi(s_t,a_t) = \sum_{t'=t}^T \gamma^{t'-t} r_{t'}$, where $r_{t^\prime}$ is the empirical reward obtained at timestep $t^\prime$ with a rollout starting with $(s_t, a_t)$ and following policy $\pi$ afterwards. $\pi_\theta$ is a policy parameterized by $\theta$.
A common improvement of the vanilla REINFORCE algorithm is to introduce a learned value baseline, $V_\phi$, to lower the variance of return estimation while maintaining unbiased. We use L2 loss to learn the value baseline: $l_{V_\phi} = ||V_\phi(s_t)-\sum_{t'=t}^T \gamma^{t'-t} r_{t'}||^2$. 
For fine-tuning large language models, it is also common to include a Kullback–Leibler (KL)-divergence regularization between the reference model and the fine-tuned model~\cite{ahmadian2024back}. This augments the final objective to:
{
\small
\begin{align}
\label{eq:reinforce_baseline_kl}
\begin{split}
\max_{\pi_\theta} \mathbb{E}_{(s_t, a_t)\sim \pi_\theta} \biggl [ & (1-\alpha) \sum_{t=0}^{T} \log \pi_\theta(a_t | s_t)  ( \tilde{Q}_{\pi_\theta}(s_t, a_t) - V_\phi(s_t)) \\
& - \alpha D_\text{KL} (\pi_\theta(\cdot | s_t) || \pi_{\text{ref}} (\cdot | s_t ) ) \biggr ]
\end{split}
\end{align}
}
where $\alpha$ controls the relative weight for the KL regularization, and $\pi_\text{ref}$ is the base reference policy. 

\noindent\textbf{Normalization of advantages}. The advantage estimate, $\tilde{A}_\pi$, is defined as $\tilde{A}_\pi(s, a) = \tilde{Q}_\pi(s,a) - V_\phi(s)$. We follow standard practice to normalize the advantage estimates across a batch of transitions: $A_\pi = (\tilde{A}_\pi - \text{mean}(\tilde{A}_\pi)) / \text{std}(\tilde{A}_\pi)$.

\subsection{PaLI-3: Vision-Language Models}
In this work, we utilize PaLI-3~\cite{chen2023palix,chen2023pali}, as the MLLM.  PaLI-3 has 5B parameters, distributed between a 2B contrastively pre-trained Vision Transformer (ViT) and a 3B fused encoder-decoder~\cite{zhai2023sigmoid}. The ViT takes images as input and produces vision embeddings. The fused encoder-decoder takes the vision embeddings and language tokens to predict language tokens in an auto-regressive fashion. PaLI-3 has shown competitive performance across various multimodal benchmarks and computational efficiency suitable for fine-tuning~\cite{chen2023pali}. 

\section{Method: MAGNIFIED}
\label{sec:method}

\begin{figure}[ht]
    \centering
    \includegraphics[width=0.8\columnwidth]{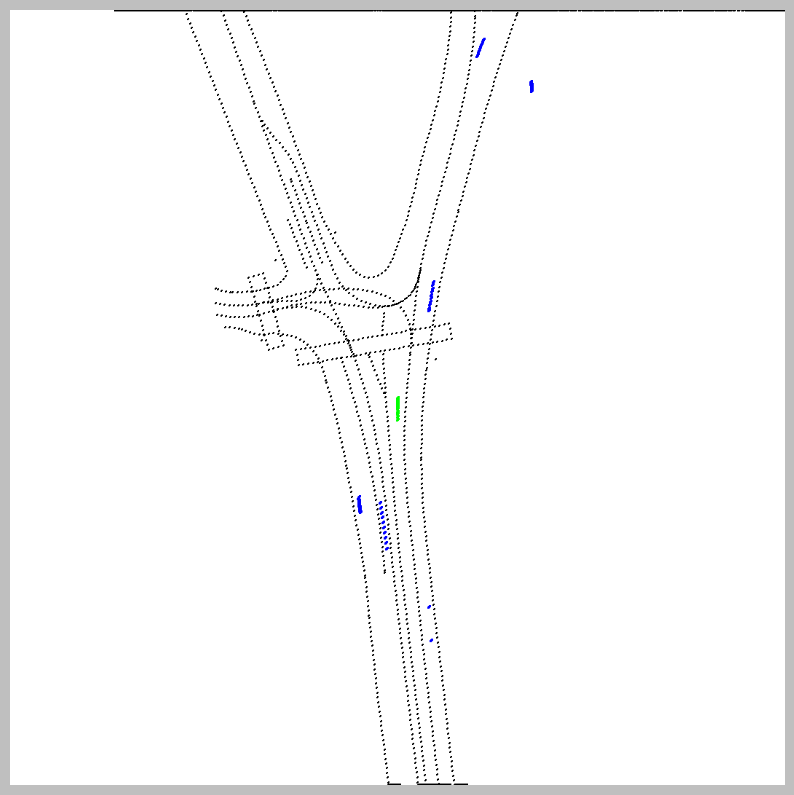}
    \caption{Example of rasterized image. The roadgraph is shown in black for better visibility. }
    \label{fig:rasterized}
    \vspace{-5pt}
\end{figure}

\begin{figure*}[t]
    \centering
    \includegraphics[width=0.98\textwidth]{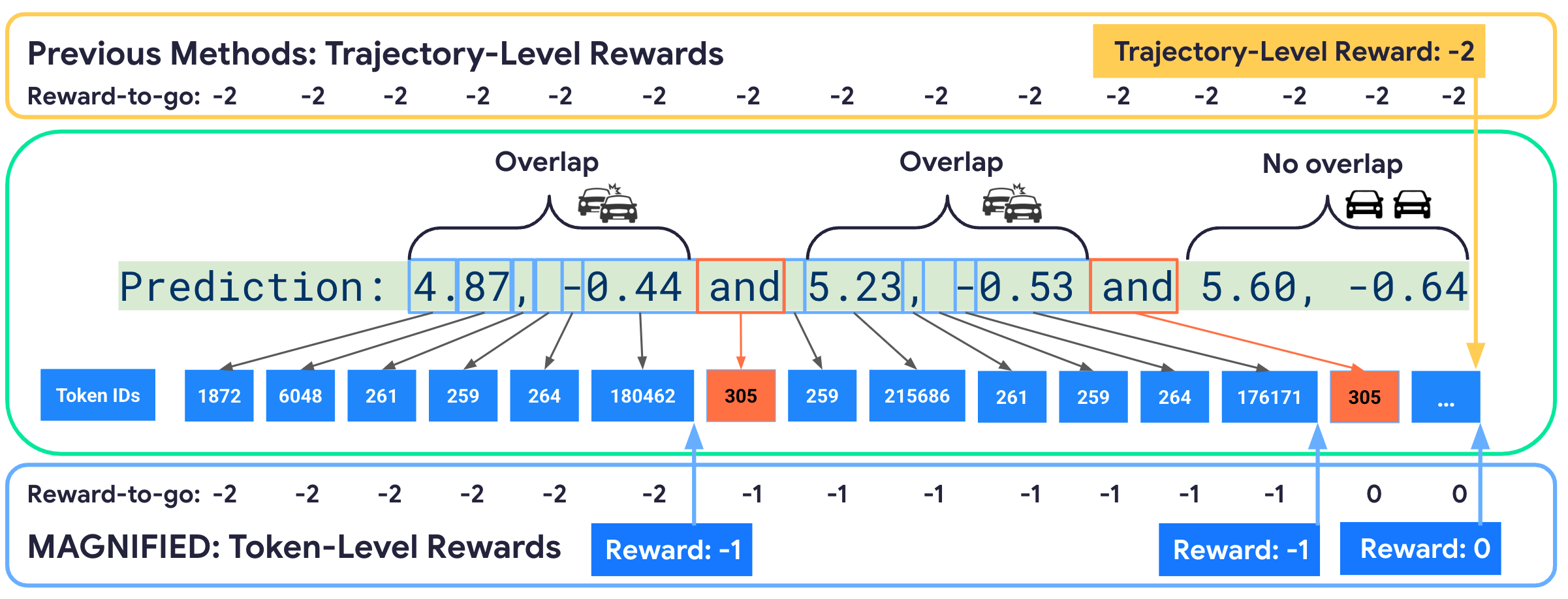}
    \caption{This figure illustrates the comparison between token-level reward and sequence-level reward. In this example, the model predicts three waypoints: (4.87, -0.44), (5.23, -0.53), and (5.60, -0.64). The corresponding token IDs are displayed in the blue boxes. Previous methods calculate the total overlap for the entire trajectory and assign a single reward at the end, resulting in a constant reward-to-go. In contrast, MAGNIFIED assigns rewards at the token level by associating each reward with the token preceding an ``and'' token (e.g., token 305). This approach captures the impacts of key tokens on the objective, facilitating more precise credit assignment. } 
    \label{fig:method}
    \vspace{-10pt}
\end{figure*}

We present MAGNIFIED, a Reinforcement Learning Fine-Tuning (RLFT) framework that enhances MLLMs for AV planning. This section outlines the problem setup, input representation, reward formulation, and the reinforcement learning used in MAGNIFIED.

\subsection{Problem Setup}
We formulate the AV planning task as a trajectory generation problem where the planner predicts the future trajectory of the ego-vehicle given the scene context, including past trajectories of other agents and the roadgraph. Trajectories are represented as a sequence of $(x, y)$ waypoints. 
In this work, we use the WOMD~\cite{ettinger2021large}, which provides rich data for planning scenarios. Each scenario involves predicting an 8-second trajectory for the ego-vehicle based on 1 second of historical information. The dataset includes information on the location, speed, and acceleration of the ego-vehicle and other agents, along with roadgraph information. The model is tasked to output the predicted trajectory as a sequence of $(x, y)$ waypoints. The data frequency is 10Hz, meaning that each input sequence consists of 10 waypoints for the past second, and the model must generate 80 waypoints.

\vspace{-5pt}
\subsection{Visual and Textual Input Representation} 
\label{subsec:input_representation}
A key challenge in MLLM planning is designing effective input modalities. We propose a novel visual and textual representation tailored for trajectory planning.
To encode the driving scene, we rasterize the roadgraph and past trajectories into a 3-channel (RGB) image with shape $[1064, 1064, 3]$. An example is shown in Fig~\ref{fig:rasterized}. Each channel conveys distinct scene elements: \textbf{Red}: represents the roadgraph, including lane boundaries and other road features. \textbf{Green}: encodes the past 1-second trajectory of the ego-vehicle. \textbf{Blue}: displays the past 1-second trajectories of other vehicles in the scene.

All coordinates are transformed to center the ego-vehicle at $(0,0)$ and its orientation facing upward. This concise representation captures the traffic flow, relative positions, and recent motion history in a form that MLLMs can intuitively interpret. We find that PaLI-3's ViT encoder effectively processes this rasterized BEV input even without fine-tuning.

In addition to the visual input, textual inputs include ego-vehicle's position, velocity, and acceleration over the past second. We also include route~\cite{gulino2024waymax}, which consists of the logged future trajectory projected to the roadgraph points. We sample 20 points from the route with a 5-point interval. The projection and sampling minimizes the information leaked about the future path. The off-the-shelf PaLI tokenizer was used to encode the text and allow MAGNIFIED to leverage the natural language knowledge prior from PaLI.

\noindent One example of the textual input is shown below:

\begin{fullwidthquote}\small\ttfamily
Answer in en: Assume I am at the coordinate 0,0.The high-level behavior attention is: go follow route:-1.71, -0.16 and 3.27, -0.13... The past trajectory under vehicle coordinate is: -2.21, 0.00 and -1.93, 0.00... The past ego velocity under vehicle coordinate is: 2.75, -0.00 and 2.61, -0.00... The past ego acceleration under vehicle coordinate is: -1.21, 0.02 and -1.33, -0.01... Other agent current locations under vehicle coordinate is: 27.81, 3.80 and -11.59, -0.61... What is my future trajectory in next 8 seconds under vehicle coordinate?
\end{fullwidthquote}

\noindent One example of the output trajectory is: 

\begin{fullwidthquote}\small\ttfamily
0.14, -0.00 and 0.27, -0.00 and 0.39, -0.00 and ...
\end{fullwidthquote}





\subsection{Token-level Rewards}
To align the MLLM-based planner with planning objectives, we propose token-level rewards, facilitating more precise credit assignment. In this work, we focus on two core planning objectives: 1) avoiding overlaps with other actors, 
and 2) avoiding driving off-road.


MAGNIFIED evaluates the cost of each predicted waypoint by checking for overlaps with other agents and whether the agent is off-road. The reward function is defined as a combination of the overlap counts and the off-road indicator by a coefficient, $w_o$:  in Equation~\ref{eq:collsion_offroad_rew}, $c_t^\text{overlap}$ denotes the number of overlapped agents at time $t$, and $\mathbb{I}_t^\text{off-road}$ denotes whether the vehicle is off-road. 
\begin{align}
r_t = -(1-w_o) c_t^\text{overlap} -w_o \mathbb{I}_t^\text{off-road}
\label{eq:collsion_offroad_rew}
\end{align}
Unlike typical RLFT in LLMs which often rely on sequence-level rewards, MAGNIFIED leverages the structure of trajectory outputs to provide dense supervision. The predicted trajectory is represented as a sequence of $(x, y)$ waypoints separated by the token ``and'', and rewards are assigned to the token preceding each ``and'' (Figure~\ref{fig:method}). This structure facilitates finer-grained credit assignment by linking tokens to planning outcomes, such as overlap or off-road violations. 

\begin{table*}[t]
\small
\centering
\resizebox{\linewidth}{!}{
\begin{tabular}{lccccc}
\toprule
\textbf{Model} & \textbf{ADE@8s$\downarrow$} & \textbf{Overlap Rate$\downarrow$} & \textbf{Overlap Count$\downarrow$} & \textbf{Off-Road Rate$\downarrow$} & \textbf{Route Progress} \\
\midrule
\multicolumn{6}{l}{\textbf{Baselines}} \\
Wayformer (re-plan)~\cite{gulino2024waymax}\textsuperscript{\dag} & N/A & 10.68\% & N/A & 7.89\% & 123.58\% \\
SFT & 1.785 & 10.10\% & 1.71 & 5.60\% & 94.9\% \\
SFT w/ blackout images & 3.179 & 18.9\% & 3.29 & 7.80\% & 105.7\% \\
SFT w/o route & 1.787 & 10.7\% & 1.78 & 6.64\% & 96.4\% \\
\midrule
\multicolumn{6}{l}{\textbf{MAGNIFIED (Ours)}} \\
$w_o=0.0$ (Overlap only) & 1.712 & 8.76\% & 1.48 & 5.88\% & 100.5\% \\
$w_o=0.25$ & \textbf{1.704} & \textbf{8.60\%} & \textbf{1.41} & 3.68\% & 103.1\% \\
$w_o=0.5$ & 1.726 & 9.04\% & 1.51 & \textbf{3.42\%} & 100.9\% \\
$w_o=0.75$ & 1.722 & 9.16\% & 1.54 & 3.56\% & 98.6\% \\
$w_o=1.0$ (Off-road only) & 1.770 & 10.06\% & 1.70 & 3.57\% & 95.4\% \\
\bottomrule
\end{tabular}}
\caption{Comparisons between baselines, SFT, and MAGNIFIED (ours) on planning metrics. \textbf{Overlap Rate} and \textbf{Off-Road Rate} are binary for each run segment and averaged over dataset. \textbf{Overlap Count} represents the averaged number of overlap instances in each run segment.
\textsuperscript{\dag} Wayformer (re-plan) uses the same route as MAGNIFIED and re-plans every 5 steps~\cite{gulino2024waymax}. 
}
\label{tab:planningresults}
\vspace{-10pt}
\end{table*}

\subsection{Policy Gradient with KL Penalty}
At the core of MAGNIFIED is a RLFT framework that leverages token-level rewards to align the MLLM with cost-aware planning objectives. We choose to utilize REINFORCE augmented with a KL-divergence penalty to optimize the planning objectives while maintaining output's trajectory structure and preventing catastrophic forgetting.

MAGNIFIED first parses the model’s output tokens as a sequence of $(x, y)$ waypoints separated by the token ``and''. The reward-to-go, $\Tilde{Q}$, is then calculated and used to compute the policy gradient  (Equation~\ref{eq:reinforce_baseline_kl}). By associating rewards with the corresponding tokens, MAGNIFIED captures the cumulative impact of tokens on the trajectory quality. While our implementation of MAGNIFIED focuses on overlap and off-road avoidance, the framework is adaptable to other planning objectives. 


\section{Results} 
\label{sec:results}

In this section, we present the experimental results on the Waymo Open Motion Dataset (WOMD)~\cite{ettinger2021large}, which includes 483{,}433 training samples and 43{,}783 evaluation samples. All experiment hyper-parameters are summarized in Table~\ref{tab:hyperparameters}. Our experiments are designed to answer the following questions:

\textbf{Q1}: Can the SFT stage take in our proposed visual and textual inputs (Sec. ~\ref{subsec:input_representation}) and enable PaLI-3 to predict structured trajectories that achieve decent imitative performance?

\textbf{Q2}: Can MAGNIFIED improve \emph{targeted planning objectives} compared to the SFT baseline?

\textbf{Q3}: Can MAGNIFIED simultaneously improve \emph{multiple planning objectives}?

\textbf{Q4}: Is token-level reward more effective than traditional sequence-level rewards?

\textbf{Q5}: Do KL regularization and other hyper-parameters impact RLFT performance?
\subsection{Benchmark Results}
We evaluate MAGNIFIED and SFT on both imitation and planning performance metrics. SFT models are trained for 100{,}000 steps with a batch size of 256. RLFT models are trained for 10{,}000 steps with a batch size of 32. Results are reported on the evaluation set.  

\noindent\textbf{Trajectory Prediction Metrics.} Table~\ref{tab:imitation_results} reports Average Displacement Error (ADE) and Final Displacement Error (FDE) at 3, 5, and 8 seconds. ADE@Ns represents the average L2 distance between the prediction and ground-truth ego-vehicle positions over the horizon of N seconds. FDE@Ns represents the final L2 distance after N seconds has elapsed. 
All baselines, including MotionLM~\cite{seff2023motionlm}, Wayformer~\cite{nayakanti2023wayformer} and EMMA~\cite{hwang2024emma}, sample multiple trajectories (24-192 samples), which are subsequently aggregated clustering into the final trajectory, whereas our approaches only generates one trajectory. Nevertheless, SFT achieves comparable imitative performance, suggesting it successfully enables PaLI-3 to process rasterized image inputs and text instructions, and generate trajectories in the structured format, establishing a strong baseline for evaluation. In particular, the significant degradation of SFT w/ blackout images (e.g., 78\% increases on ADE@8s) highlights the importance of the visual input. 

\begin{figure}[t]
    \centering
    \includegraphics[width=0.48\textwidth]{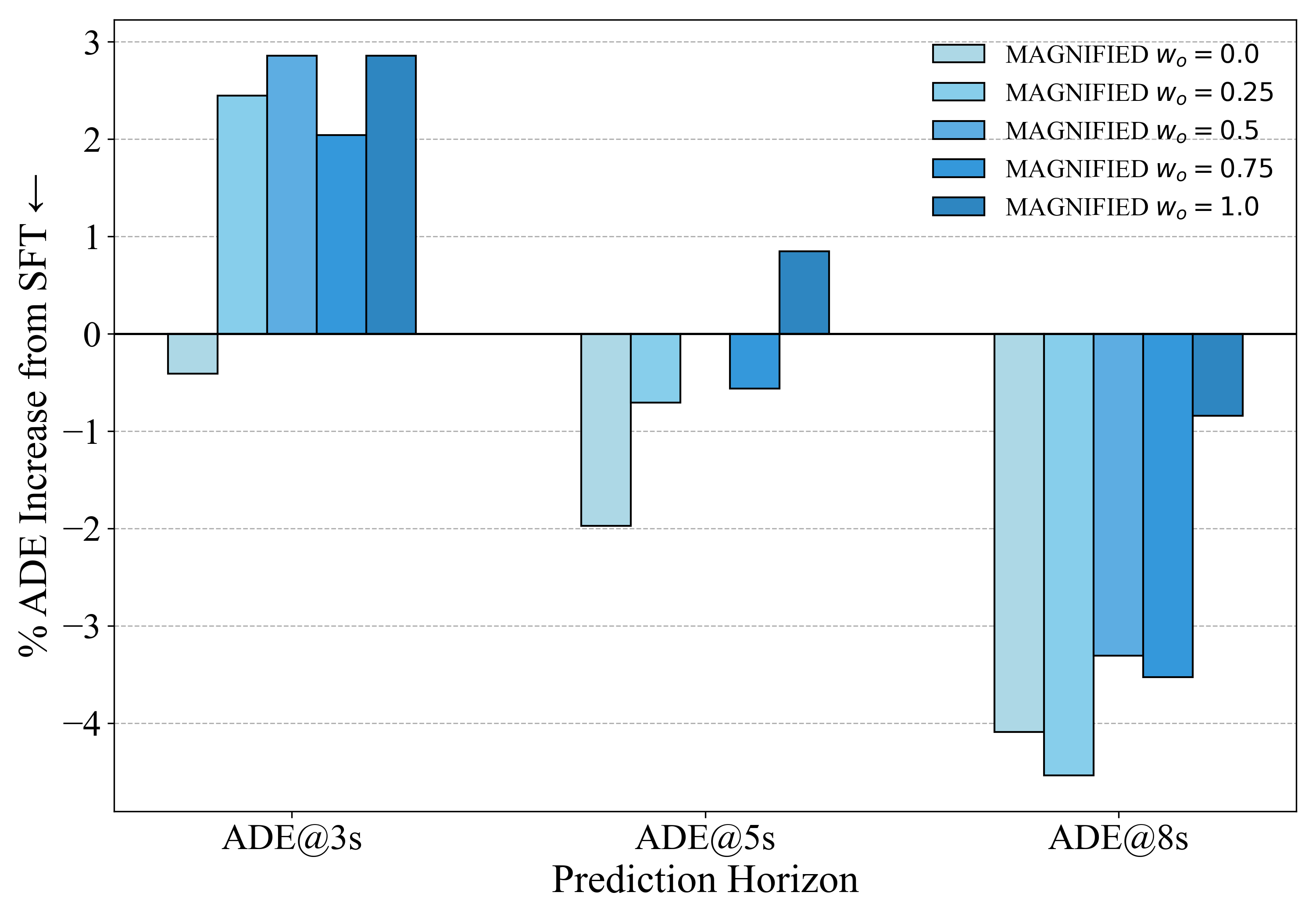}
    \caption{Percentage change in ADE at 3, 5, and 8 seconds for MAGNIFIED relative to SFT. Negative values indicate improvements over SFT. MAGNIFIED achieves better long-horizon imitation \emph{without} explicitly optimizing ADE.}
    \label{fig:ade_comparison}
    \vspace{-10pt}
\end{figure}

Surprisingly, although MAGNIFIED does not explicitly optimize ADE or FDE, it achieves better long-horizon imitative accuracy: MAGNIFIED ($w_o=0.25$) achieves a 4.5\% ADE@8s improvement and a 8.0\% FDE@8s improvement over SFT, as shown in Figure~\ref{fig:ade_comparison}. Performance at 3s and 5s remains comparable. We hypothesize that optimizing for key planning objectives captures fundamental driving intents, resulting in improved long-horizon imitative performance.

\begin{table*}[t]
\small
\centering
\resizebox{\linewidth}{!}{
\begin{tabular}{lcccccc}
\toprule
\textbf{Model} & \textbf{ADE@3s $\downarrow$} & \textbf{ADE@5s $\downarrow$} & \textbf{ADE@8s $\downarrow$} & \textbf{FDE@3s $\downarrow$} & \textbf{FDE@5s $\downarrow$} & \textbf{FDE@8s $\downarrow$} \\
\midrule
\multicolumn{7}{l}{\textbf{Baselines}\textsuperscript{*}} \\
MotionLM~\cite{seff2023motionlm}& 0.251 & 0.694 & 1.766 & N/A & N/A & N/A \\
Wayformer~\cite{nayakanti2023wayformer}& 0.250 & 0.640 & 1.517 & N/A & N/A & N/A \\
EMMA~\cite{hwang2024emma} & 0.248 & 0.681 & 1.718 & N/A & N/A & N/A \\
\midrule
\multicolumn{7}{l}{\textbf{MAGNIFIED (Ours)}} \\
Overlap only $w_o=0.0$ & 0.244 & 0.695 & 1.712 & 0.726 & 2.042 & 4.812 \\
$w_o=0.25$ & 0.251 & 0.704 & 1.704 & 0.743 & 2.047 & 4.714 \\
$w_o=0.5$ & 0.252 & 0.709 & 1.726 & 0.746 & 2.065 & 4.813 \\
$w_o=0.75$ & 0.250 & 0.705 & 1.722 & 0.740 & 2.056 & 4.823 \\
Off-road only $w_o=1.0$ & 0.252 & 0.715 & 1.770 & 0.749 & 2.101 & 5.016 \\
\midrule
\multicolumn{7}{l}{\textbf{Supervised Tuning}} \\
SFT & 0.245 & 0.709 & 1.785 & 0.735 & 2.106 & 5.124 \\
SFT w/ blackout images & 0.469 & 1.260 & 3.179 & 1.335 & 3.661 & 9.334 \\
SFT w/o route & 0.248 & 0.711 & 1.787 & 0.739 & 2.103 & 5.126 \\
\bottomrule
\end{tabular}}
\caption{Comparison between baselines, SFT, and MAGNIFIED (ours) on imitative metrics in Waymo Open Motion Dataset. * Baseline ADE results reported in ~\cite{hwang2024emma}.}
\label{tab:imitation_results}
\vspace{-10pt}
\end{table*}

\noindent\textbf{Planning Metrics.} Table~\ref{tab:planningresults} presents results for planning metrics: Overlap Rate, Overlap Count, Off-Road Rate, and Route Progress, following~\cite{gulino2024waymax}. These metrics evaluate the quality of the generated trajectory, beyond mere imitation. When optimizing only for overlap avoidance, MAGNIFIED reduces Overlap Rate from 10.10\% (SFT) to 8.76\% (a 13.3\% improvement) and Overlap Count from 1.71 to 1.48 (a 13.5\% improvement), confirming its ability to learn overlap-avoidance behavior. When optimizing only for off-road avoidance, MAGNIFIED reduces off-road rate from 5.60\% to 3.57\%, a 36.3\% reduction. These results confirm that MAGNIFIED can successfully optimize for specific planning objectives, such as overlap and offroad, while keeping the behavior neutral as shown in ADE@8s. Notably, these improvements are achieved using only 1.25\% training steps of SFT, demonstrating the sample efficiency of MAGNIFIED.

\noindent\textbf{Route Information. } To evaluate the contribution of route input, we compare ``SFT'' with ``SFT w/o route'' in Table~\ref{tab:planningresults} and Table~\ref{tab:imitation_results}. The ``SFT w/o route'' variant instead provides a coarse high-level navigation command (go straight/left/right). Although this version still produces reasonable trajectories, we observe slight improvements across all metrics when route information is included. The close performance between the two conditions suggests that the MLLM-planning paradigm is robust to degraded input information.

\subsection{Rewards Trade-off Analysis}

\begin{figure}[t]
    \centering
    \includegraphics[width=0.45\textwidth]{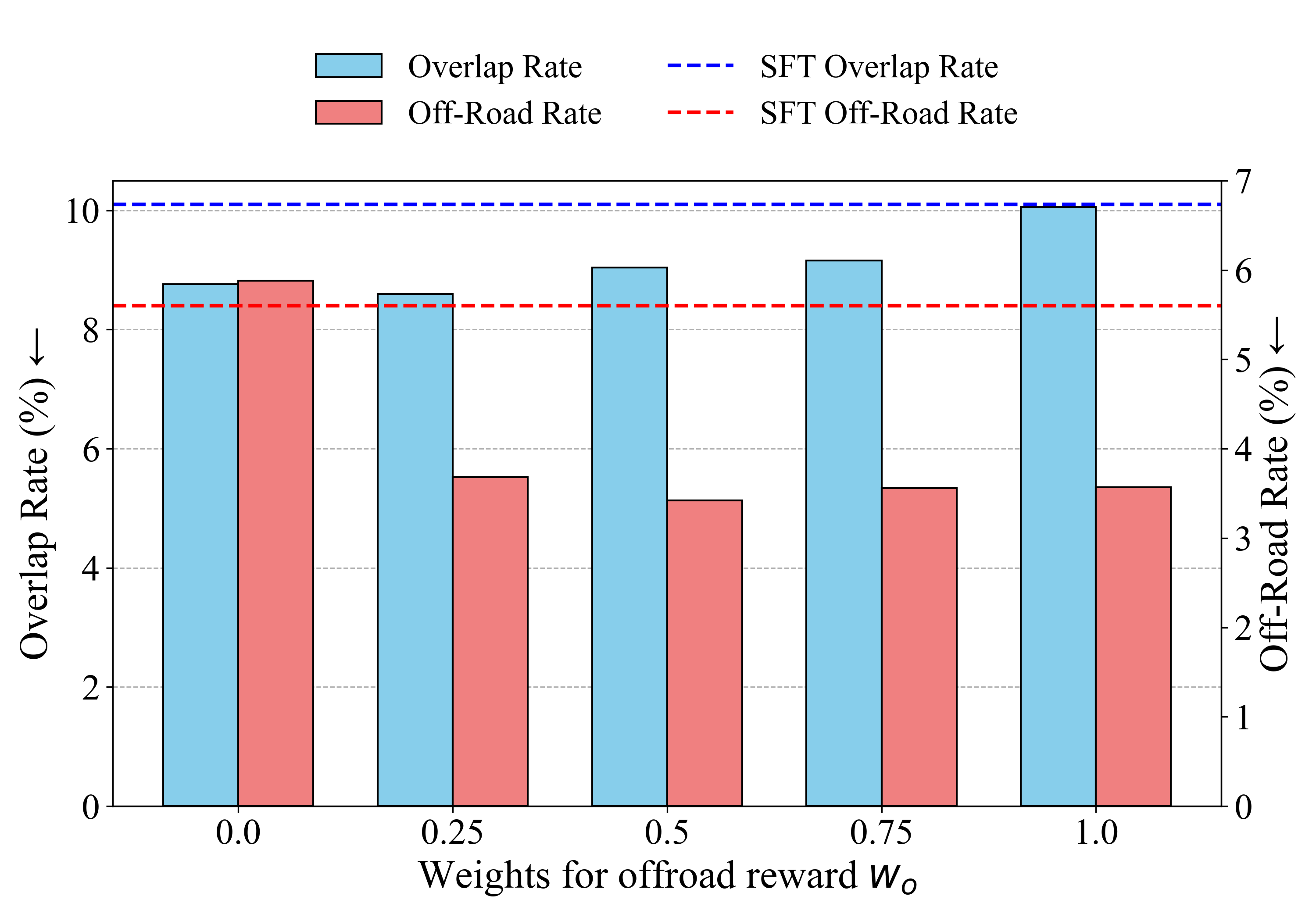}
    \caption{Effect of the off-road weight $w_o$ on planning metrics. As $w_o$ increases, overlap and off-road trade off, confirming MAGNIFIED's ability to balance multiple objectives.}
    \label{fig:reward_tradeoff}
    \vspace{-10pt}
\end{figure}

To answer \textbf{Q3}, we evaluate whether MAGNIFIED can simultaneously optimize multiple planning objectives—specifically, reducing both overlap and off-road events. We vary the reward weight $w_o$ in Equation~\ref{eq:collsion_offroad_rew}, which controls the emphasis on minimizing off-road instead of overlap. 
Figure~\ref{fig:reward_tradeoff} illustrates how varying $w_o$ affects the Overlap Rate and Off-road Rate. As $w_o$ increases, the Overlap Rate generally increases while the Off-road Rate decreases. 
On planning metrics (Table~\ref{tab:planningresults}), MAGNIFIED successfully improves both overlap and off-road metrics when using mixed-objective rewards. For example, with $w_o = 0.5$, MAGNIFIED reduces Overlap Rate from 10.10\% (SFT) to 9.04\% (a 10.5\% improvement), and Off-road Rate from 5.60\% to 3.42\% (a 38.9\% reduction). Notably, the mixed setting $w_o = 0.25$ further reduces the Overlap Rate to 8.60\% (a 14.9\% improvement) and the Off-road Rate to 3.68\% (a 34.3\% improvement). In contrast, when optimizing only for one objective, MAGNIFIED improves that metric while slightly compromising the other. For example, overlap-only reward reduces Overlap Rate but results in a slightly higher Off-road Rate of 5.88\%. 
These findings confirm that MAGNIFIED not only is able to optimize individual planning objectives, but also is able to improve multiple objectives simultaneously, all while preserving imitative performance.


\subsection{Ablation Studies}


We conduct ablations to assess the importance of design choices and answer \textbf{Q4} and \textbf{Q5}.

\begin{table}[t]
\small
\centering
\resizebox{\linewidth}{!}{
\begin{tabular}{lccc}
\toprule
\textbf{Model} & \textbf{ADE@8s} & \textbf{Overlap Rate} & \textbf{Overlap Count} \\
\midrule
$\alpha=0.0$ (w/o KL) & 1.872  & 9.76\% & 1.57  \\
\midrule
$\alpha=0.1$ & 1.712 & \textbf{8.76\%} & \textbf{1.48} \\
$\alpha=0.3$ & \textbf{1.704} & 8.79\% & 1.49  \\
$\alpha=0.5$ & 1.749 & 9.76\% & 1.63  \\
\midrule
$\alpha=0.1$, w/o Token-Rew & 1.760 & 9.92\% & 1.69  \\
\bottomrule
\end{tabular}}
\caption{MAGNIFIED ablation results. All runs use Overlap only reward ($w_o=0.0$). }
\label{tab:planningresults_ablation}
\vspace{-10pt}
\end{table}

\noindent\textbf{Token-Level Reward.} We compare token-level MAGNIFIED against a variant that uses sequence-level rewards—i.e., summing rewards across the trajectory as a single scalar reward. This ablation, reported in Table~\ref{tab:planningresults_ablation} (row “w/o Token-Rew”), isolates the contribution of our token-level reward assignment. The results suggest that token-level rewards enable more effective learning and improve planning performance compared with sequence-level rewards.

\begin{figure}[t]
    \centering
    \includegraphics[width=0.5\textwidth]{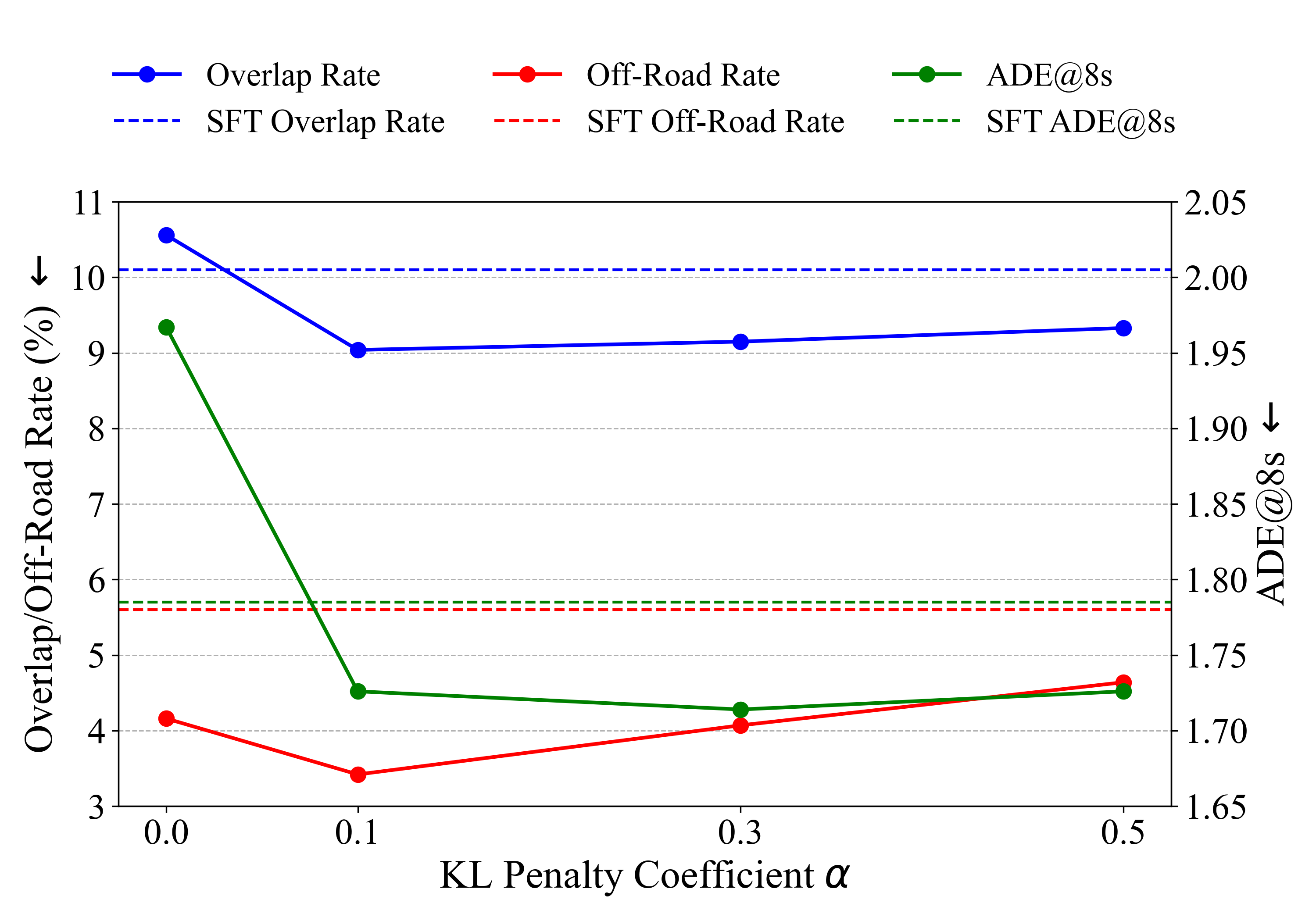}
    \caption{This figure shows the impact of KL regularization weight $\alpha$ on planning metrics ($w_o = 0.5$). Moderate regularization ($\alpha=0.1$) leads to the best performance in both Overlap Rate and Off-road Rate, while too little or too much regularization degrades performance.}
    \label{fig:ablation_kl_penalty}
    \vspace{-10pt}
\end{figure}
\noindent\textbf{RL Hyper-parameters. } We conduct experiments with different values of the KL regularization weight $\alpha\in\{0.1, 0.3, 0.5\}$, which controls the strength of penalty for divergence from the reference policy. 
We show results in Table~\ref{tab:planningresults_ablation} and Figure~\ref{fig:ablation_kl_penalty}. 
While all variants lead to improvements over the SFT baseline, we find that $\alpha = 0.1$ achieves the best performance. 
This supports the intuition that KL regularization serves as a stabilizing auxiliary term, while overly strong penalties may limit performance.

\noindent\textbf{KL Penalty. } 
To further assess the role of the KL penalty, we ablate the KL loss entirely (``w/o KL'' in Table~\ref{tab:planningresults_ablation} and $\alpha=0.0$ in Figure~\ref{fig:ablation_kl_penalty}). 
Removing the KL penalty results in degraded performance across both planning and imitation metrics, highlighting its importance in stabilizing learning and preserving imitative behaviors.


\subsection{Qualitative Analysis}
We present examples illustrating how MAGNIFIED resolves overlap instances from SFT in the supplementary video. In the Overlap Case 1, the SFT-controlled ego-vehicle moves too quickly when another vehicle merges into its lane. At $t=3s$, the ego-vehicle overlaps with the orange vehicle. MAGNIFIED, however, adjusts its speed and yields to the merging vehicle, demonstrating better planning and awareness of the merging vehicle's trajectory. More qualitative examples in the video demonstrate how MAGNIFIED effectively addresses overlap and offroad risks with RLFT on planning objectives. 

\begin{table}[t]
\centering
\begin{tabular}{cc}
\toprule
\textbf{Hyper-parameter} & \textbf{Value / Scheme} \\
\midrule
SFT learning rate & \begin{tabular}{@{}c@{}}linear warm-up from 0 to 3e-3 in 1,000 steps, \\ then delay with slope 1e-4 / $\sqrt{\text{step}/1000}$ \end{tabular} \\
\midrule
SFT batch size & 256 \\
\midrule
SFT training steps & 100,000 \\
\midrule
RLFT learning rate & \begin{tabular}{@{}c@{}}linear warm-up from 0 to 1e-4 in 2,000 steps, \\ then constant 1e-4\end{tabular} \\
\midrule
RLFT batch size & 32 \\
\midrule
RLFT training steps & 10,000 \\
\midrule
MDP temporal discount & 1.0 \\
\bottomrule
\end{tabular}
\caption{Hyper-parameters and their values.}
\label{tab:hyperparameters}
\vspace{-10pt}
\end{table}
\section{Conclusion}
In this work, we present MAGNIFIED, an RLFT framework that transforms MLLMs into cost-aware autonomous driving planners.
Our approach leverages MLLMs' semantic understanding and common-sense reasoning capabilities, and an RLFT phase integrating novel token-level rewards to directly optimize for planning objectives. Experiments on WOMD demonstrate significant overlap reduction ($>10\%$) and off-road reduction ($>38\%$) compared to baseline SFT, while maintaining or improving the imitative behaviors.

Future work could incorporate additional reward signals, enabling MAGNIFIED to address a broader range of planning objectives and real-world constraints. Another line of future work is to test MAGNIFIED beyond AV planning, such as robotics, where multi-turn RL with per-token rewards is suitable to enable long-horizon planning capabilities.

\section{Limitations}
While MAGNIFIED demonstrates strong improvements in planning metrics, several limitations remain.
First, our rewards focus only on overlap and off-road penalties, whereas real-world planning involves broader objectives such as comfort and road rules compliance.
Second, MAGNIFIED relies on a specific token structure for reward assignment, which may limit generalization to other output formats.
Third, our experiments are conducted exclusively on WOMD; evaluating transferability to other domains is left for future work. 
Finally, all evaluations of MAGNIFIED are performed open-loop -- closed-loop evaluation is important to assess compounding errors and interactive behaviors. We expect that reductions in overlap and off-road deviations at the trajectory level would translate to improved performance in closed-loop simulation. However, closed-loop effects such as distribution shift and multi-agent interactions may introduce additional challenges. 







\bibliography{reference}
\bibliographystyle{IEEEtran}

\end{document}